\documentclass{article}
\usepackage{spconf,amsmath,graphicx}
\usepackage[shortlabels]{enumitem}
\usepackage[english]{babel}
\usepackage{ifthen}
\usepackage{fancyhdr}

\title{MULTI-RESOLUTION DUAL-TREE WAVELET SCATTERING NETWORK FOR SIGNAL CLASSIFICATION}
\name{Amarjot Singh and Nick Kingsbury}
\address{Signal Processing Group, Department of Engineering, University of Cambridge, U.K.}

\begin{document}
%
\maketitle
\begin{abstract}
This paper introduces a Deep Scattering network that utilizes Dual-Tree complex wavelets to extract translation invariant representations from an input signal. The computationally efficient Dual-Tree wavelets decompose the input signal into densely spaced representations over scales. Translation invariance is introduced in the representations by applying a non-linearity over a region followed by averaging. The discriminatory information in the densely spaced, locally smooth, signal representations aids the learning of the classifier. The proposed network is shown to outperform Mallat's ScatterNet~\cite{Jbruna2013} on four datasets with different modalities on classification accuracy.
\end{abstract}
\section{Introduction}
\label{sec:intro}
Signal classification is a difficult problem due to the considerable translation, rotation and scale variations that can hinder the classifier's ability to measure signal similarity~\cite{BAJ}. Deep Convolutional Neural Networks (CNNs)~\cite{LeCun1998} have been widely used to eliminate the above-mentioned variabilities and learn invariant as well as discriminative signal representations by using successive kernel operations (linear filters, pooling, and non-linearity). Despite their success, the optimal configuration of these networks is not well understood because of the cascaded nonlinearities. 

Scattering convolution network proposed by S. Mallat in \cite{Jbruna2013} provided a mathematical framework to incorporate geometric signal priors to extract discriminative and invariant signal representations. Invariance is introduced in the representations by filtering the input signal with a cascade of multiscale and multidirectional complex Morlet wavelets followed by pointwise nonlinear modulus and local averaging. The high frequencies lost due to averaging are recovered at the later layers using cascaded wavelet transformations with non-linearities, justifying the need for a multilayer network. The above class of networks has been widely used in numerous applications as such as object classification~\cite{Oyallon2015}, audio processing~\cite{mlsp}, scene understanding~\cite{scene}, biology~\cite{biology} etc.

This paper proposes an improved Deep Scattering architecture that uses Dual-Tree Complex Wavelet Transform (DTCWT)~\cite{Kingsbury1998} to decompose a \textit{multi-resolution input signal} into translation invariant signal representations. The input signal is first decomposed into multi-resolution signal representations that are densely spaced on the scale domain. Translation invariance is then introduced within each representation by applying a non-linearity over a region followed by local averaging. Next, a log non-linearity is used to separate the multiplicative low-frequency illumination components within the representations. Finally, a Support Vector Machine (SVM) is used to create discrimination between different signal classes by learning weights that best summarize the regularities (common coefficients) in the training data and simultaneously ignore the coefficients arising due to the irregularities~\cite{Joachims}.   

 \begin{figure*}[t!] 
\centering    
\includegraphics[scale = 0.41]{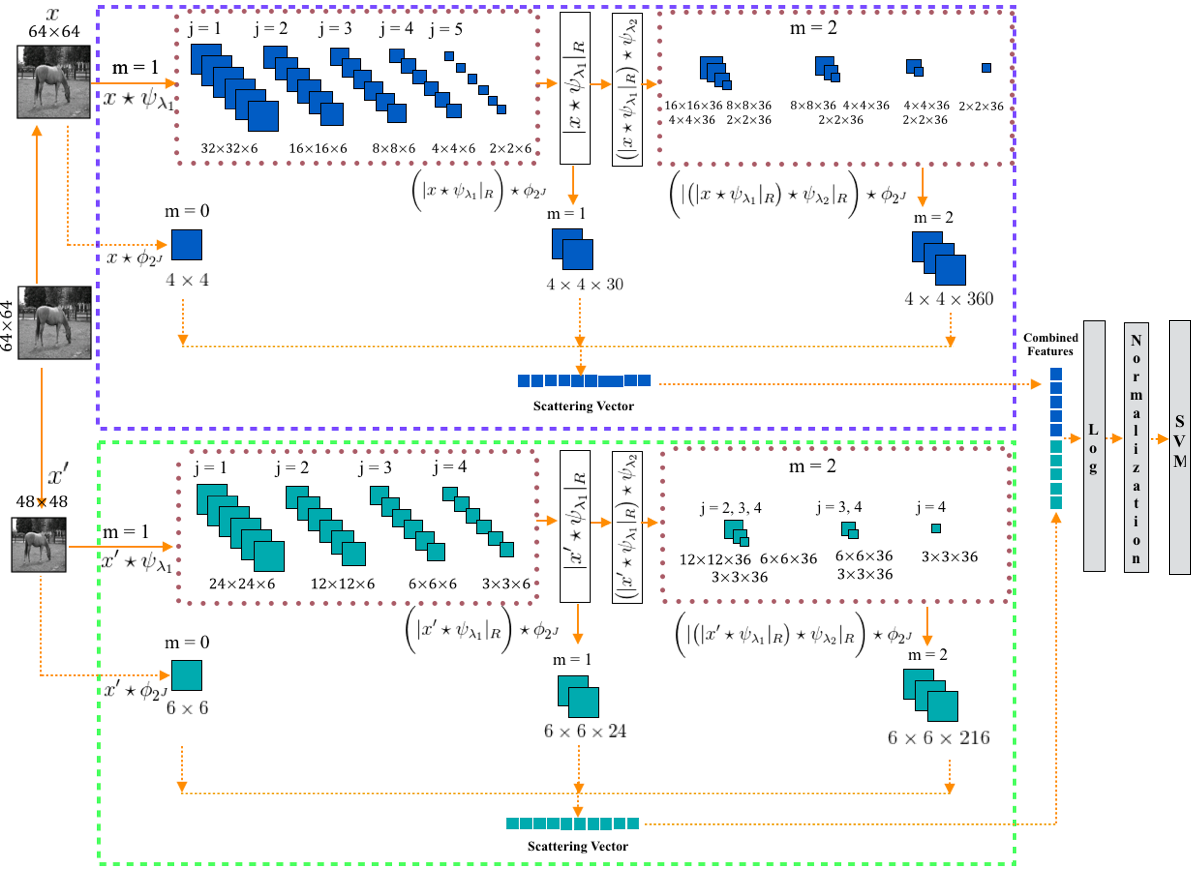}
\caption[]{\small{{Illustration shows the input image ($x$) of size $64 \times 64$ resized to images of resolution, $x$ ($64 \times 64$ ) and $x'$ ($48 \times 48$) respectively. Image representations at m = 1 are obtained using DTCWT filters at different scales and 6 orientations. Then, L2 region non-linearity is applied on the representations to obtain the regular envelope followed by local averaging to extract the translation invariant coefficients. The information lost due to averaging are recovered by cascaded wavelet filtering at the second layer. Translation invariance is introduced in the recovered frequencies using L2 region non-linearity and local averaging. }}}
\end{figure*} 

The main contributions of the paper and their reasoning are explained below:
\vspace{-1mm}
\begin{itemize}
\item \textit{Multi-scale Input Signal}: The input signal is decomposed into representations that are densely spaced on the scale space using a DTCWT based decimated pyramid of complex values~\cite{Anderson2005}. The multi-scale representations have redundant local regularities that allows the SVM to optimally learn weights that learn discriminatory features (edges between two objects within an image) from fine scale representations while non-discriminator features like the middle of the objects from the coarse features~\cite{chan}. 
\vspace{-1mm}
\item \textit{DTCWT Filter Bank}: The proposed network uses DTCWT bank for filtering as opposed to Morlet wavelets \cite{Jbruna2013} due to its perfect reconstruction properties~\cite{Kingsbury1998}. Perfect reconstruction property allows the DTCWT filter to extract features without any aliasing.
\vspace{-1mm}
\item \textit{Region Non-Linearity}: The extracted representations are cascaded by a \textit{region non-linearity} as opposed to a point non-linearity and then followed by local averaging to produce a regional translation invariant representation. The region non-linearity selects the dominant feature within the region while simultaneously suppressing features with lower magnitudes leading to invariance similar to the max operator in CNNs.

\end{itemize}

The performance of the proposed network is tested on four popular datasets selected from different modalities. Such diversity of data sets is crucial to verify the generalization of the proposed network to a large variety of problems. 

The paper is divided into the following sections. Section 2 describes the proposed Deep Multi-Resolution DTCWT Scattering Architecture while Section 3 presents the step by step experimental results leading to signal classification. Section 4 draws conclusions about the experimental results.

\section{Multi-Resolution DTCWT ScatterNet}
\label{sec:typestyle}
In this section, the mathematical formulation of the proposed Multi-Resolution DTCWT Scattering Network is presented for 2 layers that decompose a two-dimensional input signal $x$ into translation invariant representations that are densely spaced over scales. However, the formulation can be easily extended to the input signal of other dimensions and network with deeper layers.  

First, the multi-resolution representation of the signal $x$ is obtained at interleaved scale (s)~\cite{Anderson2005} using a DTCWT based decimated pyramid of complex values as shown below:
\begin{equation}
[x,x'] = decimated (x)
\end{equation}
The proposed scattering network is applied at each resolution ($x, x'$) to produce a translation invariant representation at multiple layers. For the sake of simplicity, the derivation is presented only for the input signal $x$.  

The proposed architecture is realized by arranging DTCWT filters with numerous scales (and orientations (for only 2D signal)) in multiple layers to extract stable and informative signal representations.~DTCWT is an advantageous filtering choice due to its perfect reconstruction properties. It allows the extraction of features without any aliasing. DTCWT wavelets are represented by $\psi$ (with a real $\psi^a $ and imaginary $\psi^b$ group). The complex band-pass filter $\psi$ is decomposed into real and imaginary groups as shown:
\begin{align}
\psi (t) = \psi^{a} (t) + \iota \psi^{b} (t) \quad t = (t_{1}, t_{2})
\end{align}
The signal $x$ is filtered at the first layer ($L_{1}$) using a family of DTCWT wavelets $\psi_{\lambda_{1} } (t)$ at different scales and orientations ($\lambda_{1}$), formulated as:
\begin{equation}
x\star \psi_{\lambda } (t) = x\star\psi_{\lambda }^{a} (t) + \iota x\star\psi_{\lambda }^{b} (t) 
\end{equation}
The wavelet transform response commutes with translations, and is therefore not translation invariant. To build a translation invariant representation, a $L_{2}$ smooth non-linearity is first applied over all overlapping regions of size $R$ ({$R \times R$} for 2D signal) in feature output, obtained at a particular scale (and six orientation ($\theta$) (for 2D signal)). The non-linearity applied to one of the above-mentioned regions is shown below:
\begin{equation}
|x\star \psi_{\lambda_{1}}|_{R} = \sqrt{|G_{real}\star \psi_{\lambda_{1} }^{a} (t)|^2 + |G_{imag}\star \psi_{\lambda_{1}} ^{b} (t)|^2}  
\end{equation}
where $R$ is the size of the region and $G$ is a group of $R$ ($R \times R$ for 2D signal) complex scattering coefficients. $L_{2}$ is a non-expansive non-linearity that makes it stable to additive noise and deformations~\cite{Jbruna2013}. The region non-linearity selects the dominant feature in the region while simultaneously suppressing features with lower magnitudes. This creates translation invariance in a larger region similar to the max operator in CNNs. The scattering coefficients obtained after applying the region non-linearity to the outputs of every wavelet scales is given by $|x\star \psi_{\lambda_{1}}|_{R}$.

Next, the desired translation invariant representation are obtained at the first layer ($L_{1}$) by applying a local average on $|x\star \psi_{\lambda_{1}}|_{R}$, as shown below: 
\begin{equation}
(L_{1})_{R} = \bigg(|x\star \psi_{\lambda_{1}}|_{R}\bigg) \star \phi_{2^J} 
\end{equation}
The high frequencies coefficients lost by the averaging operator are recovered at the second layer ($L_{2}$) by calculating the wavelet coefficients of $|x \star \psi_{\lambda_{1}}|_{R}$ by the wavelet at scale and orientation, $\lambda_{2}$, given as $(|x \star \psi_{\lambda_{1}}|_{R})\star \psi_{\lambda_{2}}(t)$~\cite{Jbruna2013}. 

The features extracted at the first layer ($L_{1}$) are filtered with the DTCWT filter at coarser scales ($\lambda_{2}$) to recover the high frequency components at the second layer ($L_{2}$). The recovered frequencies are converted into translation invariant representations by again taking a local average as shown:
\begin{equation}
(L_{2})_{R} = \bigg(|\big(|x \star \psi_{\lambda_{1}}|_{R}\big)\star \psi_{\lambda_{2}}|_{R}\bigg)\star \phi_{2^J}  
\end{equation}
The scattering coefficients $S_{J}x$ for the network at different scales and orientations for two layers at a path $p$ can be obtained using the following:
\begin{equation}
S_{J}x[p] = \begin{pmatrix}
x \star \phi_{2^J} \\
\bigg(|x\star \psi_{\lambda_{1}}|_{R}\bigg) \star \phi_{2^J} \\
\bigg(|\big(|x \star \psi_{\lambda_{1}}|_{R}\big)\star \psi_{\lambda_{2}}|_{R}\bigg)\star \phi_{2^J} 
\end{pmatrix}_{\lambda = (2,3,4)}
\end{equation}  
A logarithm non-linearity proposed by Oyallon et al.~\cite{Oyallon2015}, is applied to the scattering coefficients in order to transform the low-frequency multiplicative components that arise due to illuminations into additive components. These additive coefficients can now be ignored as noise by the classifier. The logarithm applied to the scattering coefficients ($Sx$) extracted from a dataset with $M$ training signals, where the coefficients computed from a single signal has $N$ dimensions, is given by:
\begin{equation}
\Phi^{M\times N} = \log(Sx^{M\times N}+k)
\end{equation}
where $Sx$ are the scattering coefficients and $k$ is (small) constant added to reduce the effect of noise magnification at small signal levels. The value of the constant $k$ ($1e^{-6}$) used in~\cite{Oyallon2015} is duplicated for all the experiments in this paper. 

\section{Overview of Results}
\label{headings}
Experiments are conducted on four real-world datasets selected from the image, audio, biology and material modalities to evaluate the performance of the proposed DTCWT multi-resolution scattering network. In order to test the generalization of the proposed network to different problems, a large mixture of data sets from different domains with various sizes and dimensionality are used for experimentation. Please see Table 1 for a detailed description of the datasets used in our experiments. 

The proposed multi-resolution DTCWT scattering network is applied on each dataset to extract translation invariant multi-scale representations that are further used for classification. Scattering coefficients in the case of the two-dimensional signals for all the experiments is computed at six orientations (15$^\circ$, 45$^\circ$, 75$^\circ$, 105$^\circ$, 135$^\circ$, 165$^\circ$). The discrimination between the signal classes is achieved using a Gaussian SVM. Before the SVM is trained on the training set, each feature is standardized by the mean and standard deviation of the training dataset.

The test set generalization error of the proposed Deep DTCWT multi-resolution scattering network is reported on each Dataset (Table. 1) and compared with the scattering network proposed by Mallat et al~\cite{Jbruna2013}. In addition, this error for the proposed network is also compared with the recently proposed machine learning approaches used for classification of the above-mentioned data sets. Only those approaches are considered that don't augment the data sets and apply their algorithm only on the unaltered input signal.
\vspace{-1mm}
\begin{table*}[!t]
\centering
\caption{Classification error (\%) on different datasets for each component of the proposed network. DTCWT ScatNet: DSCAT, DTCWT ScatNet + Pooling: DSCATP, Multi-Resolution DTCWT ScatNet: MDSCAT, Multi-Resolution DTCWT ScatNet + Pooling: MDSCATP. The left result in $/$ is without log non-linearity applied while the right is with log applied ($No Log/Log$).}
\label{components}
\begin{tabular}{c|ccccc}
\hline
 Dataset & DSCAT & DSCATP & MDSCAT & MDSCATP \\
 \hline
 USPS & 3.31 / 3.38 & 3.24 / 3.33  & 2.89 / 3.11  & \textbf{2.56} / 2.84\\
 Isolet & 5.14 / 5.3 & 5.10 / 5.36  &  4.75 / 5.02  & \textbf{4.14} / 4.88\\
 Yeast & 41.65 / 41.17 & 45.86 / 45.85  & 37.04 / \textbf{34.62}  & 39.77 / 39.04\\
 Glass & 31.78 / 29.16 & 31.77 / 33.68  &  27.82 / \textbf{24.32} & 30.05 / 26.06 \\

\end{tabular}
\end{table*}

\vspace{-1mm}

\subsection{US Postal Service Dataset} 
The US postal service dataset consists of two-dimensional structured grayscale image signals with 7291 training observations and 2007 test observations~\cite{hl}. This dataset was generated by scanning the handwritten digits from envelopes by the U.S. Postal Service. The recorded images are de-slanted and size normalized to 16 x 16 (256) pixels images in the dataset. The objective is to differentiate between 10 different digits between 0 and 9.  

The proposed scattering network extracts the input signal at 6 resolution (s) (1, 0.85, 0.70, 0.6, 0.5, 0.35) and then extracts scattering coefficients from each resolution at 3 DTCWT scales (J) and 6 orientations ($\theta$). The region non-linearity is applied on overlapping regions (R) of size 2$\times$ 2. A cost value (c) of 5 was selected for the linear SVM. All the parameters
are selected using 5-fold cross validation. As noted from Table.~\ref{components}, the proposed network with region non-linearity and without log non-linearity results in the lowest classification error of 2.54\%. This increase in error due to the log non-linearity is explained in the previous section. The classification error of the proposed architecture is also compared to ScatterNet~\cite{Jbruna2013} and Fuzzy Integral Combination algorithm~\cite{LeCun1998} as presented in Table.~\ref{USPS}. The proposed architecture outperforms both the algorithms. 
\begin{table}[!h]
\centering
\caption{Classification error (\%) comparison on USPS}
\label{USPS}
\begin{tabular}{c|cccc}
\hline
 Dataset & Proposed & ScatNet~\cite{Jbruna2013} & FIC~\cite{Hadjadji}\\
 \hline
USPS & \textbf{2.54} & 2.6 & 5.43  \\
\end{tabular}
\end{table}
\vspace{-2mm}
\subsection{The UCI Isolet Dataset} 
The Isolet dataset comprises of one-dimensional audio signals collected from 150 speakers uttering all characters in the English alphabet twice. Each speaker contributed 52 training examples with a total of 7797 recordings~\cite{Newman}. The recordings are represented with 617 attributes such as spectral coefficients, contour, sonorant and post-sonorant are provided to classify letter utterance. 

The proposed scattering network decomposes the input signal at 4 resolution (s) (1, 0.70, 0.5, 0.35). The translation invariant features are extracted at 6 scattering DTCWT wavelet scales (J) for the input signal at every resolution. Regions (R) of size 1$\times$4 is chosen for the application of the region non-linearity.  A cost value (c) of 15 was chosen for the linear SVM. Again, the parameters are selected using 5-fold cross validation. The generalization error is reported on 10-fold cross validation for this dataset. Table.~\ref{components} shows that the multi-resolution scattering architecture with region non-linearity and log non-linearity produces the lowest classification error of 4.14\%. The proposed method outperformed ScatterNet~\cite{Jbruna2013} but was unable to surpass the performance of Extreme entropy machines~\cite{Tabor} as shown in Table~\ref{ISOLET}. 
\vspace{-3mm}
\begin{table}[!h]
\centering
\caption{Classification error (\%) comparison on Isolet}
\label{ISOLET}
\begin{tabular}{c|cccc}
\hline
 Dataset & Proposed & ScatNet~\cite{Jbruna2013} & EEM~\cite{Tabor} \\
 \hline
 Isolet & 4.14 & 5.78 & \textbf{2.70} \\

\end{tabular}
\end{table}
\vspace{-4mm}
\subsection{The UCI Yeast Dataset}
This is a highly imbalanced one-dimensional signal dataset that consists of 1484 yeast proteins with 10 cellular binding sites~\cite{Newman}. Each binding site is described with 8 attributes. The aim is to classify the most probable cellular localization site of the proteins. 

The proposed scattering network decomposes the input signal at 2 resolution (s) (1, 0.70). The translation invariant features are extracted at 2 scattering DTCWT wavelet scales (J) for the input signal at every resolution. The Region (R) size of 1$\times$2 and a cost value (c) of 15 is chosen using 5-fold cross validation. The generalization error was reported on 10-fold cross validation for this dataset. Table.~\ref{components} shows that the multi-resolution scattering architecture with region non-linearity and log non-linearity produces the lowest classification error of 35.02\%. The proposed method outperformed ScatterNet~\cite{Jbruna2013} but was unable to outrank the instance selection genetic algorithm~\cite{chen} as shown in Table~\ref{yeast}. 
\vspace{-3mm}
\begin{table}[!h]
\centering
\caption{Classification error (\%) comparison on Yeast}
\label{yeast}
\begin{tabular}{c|cccc}
\hline
 Dataset & Proposed & ScatNet~\cite{Jbruna2013} & IS~\cite{chen} \\
 \hline
 Yeast & 35.02 & 35.89 & \textbf{33.0} \\

\end{tabular}
\end{table}
\subsection{The UCI Glass Dataset}    
This dataset consists of 214 one-dimensional signals that describe six types of glass based on 9 chemical fractions of the oxide content~\cite{Newman}. This dataset was motivated by a criminological investigation where the correct classification of glass left on the crime scene could be used for evidence. Hence, the aim is to classify between different types of glass. 

The proposed scattering network uses the same parameters as mentioned in Section. 3.3 for feature extraction. The generalization error was reported on 10-fold cross validation for this dataset. Table.~\ref{components} shows that the multi-resolution scattering architecture with region non-linearity and log non-linearity produces the lowest classification error of 24.32\%. The proposed method outperformed ScatterNet~\cite{Jbruna2013} and Kernelized Vector Quantization~\cite{villmann} as shown in Table~\ref{glass}. 
\vspace{-4mm}
\begin{table}[!h]
\centering
\caption{Classification error (\%) comparison on Glass}
\label{glass}
\begin{tabular}{c|cccc}
\hline
 Dataset & Proposed & ScatNet~\cite{Jbruna2013} & KVQ~\cite{villmann}\\
 \hline
 Glass & \textbf{24.32} & 28.86 & 31.6 \\
\end{tabular}
\end{table} 
\vspace{-4mm}
\section{Conclusion}
The paper proposes a ScatterNet that extracts regionally translation invariant features from an input signal that are equally spaced over the scale space. The proposed algorithm was tested on four datasets. It outperformed Mallat's ScatterNet on all the datasets while was able to outperform the learning based algorithms only on two datasets. Hence, it is necessary to take learning into account. The proposed scattering network can then provide the first two layers of such learning networks. It eliminates translation variability, which can help in learning the next layers. In addition, this network can replace simpler low-level features such as SIFT vectors.

\bibliographystyle{IEEEbib}
\bibliography{refs}

\begin{thebibliography}{10}

\bibitem{Jbruna2013}
J.~Bruna and S.~Mallat,
\newblock ``Invariant scattering convolution networks,''
\newblock {\em IEEE Transaction on Pattern Analysis and Machine Intelligence},
  vol. 35, pp. 1872 --1886, 2013.

\bibitem{BAJ}
B.~Scholkopf and A.J. Smola,
\newblock ``Learning with kernels,''
\newblock {\em MIT Press}, 2002.

\bibitem{LeCun1998}
Y.~LeCun, L.~Bottou, Y.~Bengio, and P.~Haffner,
\newblock ``Gradient-based learning applied to document recognition,''
\newblock {\em Proceedings of the IEEE}, vol. 86, no. 11, pp. 2278--2324, 1998.

\bibitem{Oyallon2015}
E.~Oyallon and S.~Mallat,
\newblock ``Deep roto-translation scattering for object classification,''
\newblock {\em IEEE CVPR}, pp. 2865--2873, 2015.

\bibitem{mlsp}
J.~Andén, V.~Lostanlen, and S.~Mallat,
\newblock ``Joint time-frequency scattering for audio classification,''
\newblock {\em Proceedings of IEEE MLSP Workshop}, 2015.

\bibitem{scene}
S.~Nadella, A.~Singh, and SN~Omkar,
\newblock ``Aerial scene understanding using deep wavelet scattering network
  and conditional random field,''
\newblock {\em in: Hua G., Jégou H. (eds) Computer Vision \textemdash~
  European Conference on Computer Vision (ECCV) Workshops}, vol. 9913, pp.
  205--214, 2016.

\bibitem{biology}
V.~Milišić and S.~Mallat,
\newblock ``Mathematical modeling of lymphocytes selection in the germinal
  center,''
\newblock {\em Journal of Mathematical Biology}, 2016.

\bibitem{Kingsbury1998}
N.G. Kingsbury,
\newblock ``Complex wavelets for shift invariant analysis and filtering of
  signals,''
\newblock {\em Applied and computational harmonic analysis}, vol. 10, pp.
  234--253, 2001.

\bibitem{Joachims}
T.~Joachims,
\newblock ``Optimizing search engines using clickthrough data,''
\newblock {\em 8th ACM SIGKDD international conference on Knowledge discovery
  and data mining}, 2002.

\bibitem{Anderson2005}
R.~Anderson, N.G. Kingsbury, and J.~Fauqueur.,
\newblock ``Determining multi-scale image feature angles from complex wavelet
  phases,''
\newblock {\em In Proceedings of the Second ICIAR}, pp. 490--498, 2005.

\bibitem{chan}
J.C. Chan, H.~Ma, and T.K. Saha,
\newblock ``Partial discharge pattern recognition using multiscale feature
  extraction and support vector machine,''
\newblock {\em 2013 IEEE Power and Energy Society General Meeting}, 2013.

\bibitem{hl}
J.J. Hull,
\newblock ``A database for handwritten text recognition research,''
\newblock {\em IEEE Transaction on Pattern Analysis and Machine Intelligence},
  vol. 16, no. 5, pp. 550--554, 1994.

\bibitem{Hadjadji}
B.~Hadjadji, Y.~Chibani, and H.~Nemmour,
\newblock ``Fuzzy integral combination of one-class classifiers designed for
  multi-class classification,''
\newblock {\em Image Analysis and Recognition, LNCS}, vol. 8814, pp. 320--328,
  2014.

\bibitem{Newman}
D.~Newman, S.~Hettich, C.~Blake, and C.~Merz,
\newblock ``\uppercase{UCI} repository of machine learning databases,''
\newblock {\em http://www.ics.uci.edu/∼mlearn/MLRepository.html}.

\bibitem{Tabor}
W.M. Czarnecki and J.~Tabor,
\newblock ``Extreme entropy machines: robust information theoretic
  classification,''
\newblock {\em Pattern Analysis and Applications}, pp. 1--18, 2015.

\bibitem{chen}
Z.Y.~Chen et~al.,
\newblock ``Instance selection by genetic-based biological algorithm,''
\newblock {\em Soft Computing}, vol. 19, no. 8, pp. 1--18, 2015.

\bibitem{villmann}
T.~Villmann, S.~Haase, and M.~Kaden,
\newblock ``Kernelized vector quantization in gradient-descent learning,''
\newblock {\em Neurocomputing}, vol. 147, pp. 83–95, 2015.

\end{thebibliography}

\end{document}